\documentclass{article} 

\usepackage{nfam2025_workshop}
\usepackage{times}

\usepackage{amsmath,amsfonts,bm}

\def\eqref#1{equation~\ref{#1}}

\def\1{\bm{1}}

\DeclareMathAlphabet{\mathsfit}{\encodingdefault}{\sfdefault}{m}{sl}
\SetMathAlphabet{\mathsfit}{bold}{\encodingdefault}{\sfdefault}{bx}{n}

\usepackage[framemethod=TikZ]{mdframed} 
\usepackage{mdframed}
\usepackage{hyperref}
 \usepackage{lipsum}
\usepackage{url}
\usepackage{amsmath}
\usepackage{amsthm}
\theoremstyle{plain}

\usepackage{mdframed}
\definecolor{mycyan}{RGB}{230, 255, 255}
\mdfsetup{
    backgroundcolor=mycyan,
    linewidth=0pt,
}
\newmdtheoremenv[linewidth=0pt,innerleftmargin=4pt,innerrightmargin=4pt]{definition}{Definition}
\newmdtheoremenv[linewidth=0pt,innerleftmargin=4pt,innerrightmargin=4pt]{proposition}{Proposition}
\newmdtheoremenv[linewidth=0pt,innerleftmargin=0pt,innerrightmargin=0pt,backgroundcolor=examplecolor]{example}{Example}

\title{Modern Hopfield Networks \\ with Continuous-Time Memories}

\author{Saul Santos~\textsuperscript{1,2}, António Farinhas~\textsuperscript{1,2}, Daniel C. McNamee~\textsuperscript{3}, André F. T. Martins~\textsuperscript{1,2,4,5}\\
\textsuperscript{1}Instituto de Telecomunicações, ~
\textsuperscript{2}Instituto Superior Técnico, Universidade de  Lisboa, \\
\textsuperscript{3}Champalimaud Research, 
~ \textsuperscript{4}Unbabel , 
~ \textsuperscript{5}ELLIS Unit Lisbon\\
\fontsize{8}{10}\texttt{\{saul.r.santos,antonio.farinhas,andre.t.martins\}@tecnico.ulisboa.pt},\\
\fontsize{8}{10}\texttt{daniel.mcnamee@research.fchampalimaud.org}\\
}

\iclrfinalcopy 
\begin{document}

\maketitle

\begin{abstract}
Recent research has established a connection between modern Hopfield networks (HNs) and transformer attention heads, with guarantees of exponential storage capacity. However, these models still face challenges scaling storage efficiently.
Inspired by psychological theories of continuous neural resource allocation in working memory, we propose an approach that compresses large discrete Hopfield memories into smaller, \textit{continuous-time} memories. Leveraging continuous attention, our new energy function modifies the update rule of HNs, replacing the traditional softmax-based probability mass function with a probability \emph{density} over the continuous memory. This formulation aligns with modern perspectives on human executive function, offering a principled link between attractor dynamics in working memory and resource-efficient memory allocation.
Our framework maintains competitive performance with HNs while leveraging a compressed memory, reducing computational costs across synthetic and video datasets.
\end{abstract}

\section{Introduction}
Hopfield networks \citep[HNs]{hopfield1982neural} are biologically inspired recurrent models that retrieve complete memories from partial cues through attractor dynamics, simulating episodic memory recall in humans and animals \citep{tyulmankov2021biological,whittington2021relating}. Classical HNs \citep{hopfield1982neural} store memories as fixed-point attractors, with storage capacity scaling linearly with the number of features. Recent research on \textbf{modern Hopfield networks} has introduced stronger nonlinearities in the update rule, enabling super-linear and even exponential storage capacity \citep{krotov2016dense,  ramsauer2020hopfield,hu2023sparse, santos2024sparse, santos2024hopfieldfenchelyoungnetworksunifiedframework}. Despite these improvements, efficiency remains a challenge. While recent work has reduced retrieval complexity by reframing memory retrieval as a regression problem \citep{hu2024nonparametric}, and \citet{hoover2024dense} recovered the compressed memory versions of classical HNs for dense associative models, it is still an open problem to increase the efficiency of Hopfield networks through \textbf{memory compression}---\emph{how can we store information in a more compact form without sacrificing retrieval performance?}

Recent psychology research has characterized the human working memory processing based on \textbf{continuous neural resource allocation} \citep{ma2014changing}. This theory posits that humans dynamically allocate neural activity across a set of stimuli or events to optimize their collective storage in a compressed format \citep{bays2008dynamic,TomicBays}. We suggest that this mechanism represents a continuous form of attention and contrasts sharply with traditional theories based on discrete memory ``slots'' \citep{miller1956magical} akin to discrete attention. Despite its empirical success in explaining human cognitive performance, this theory lacks a detailed recurrent neural network (RNN) implementation explaining memory-related population activity dynamics in prefrontal cortex (PFC) \citep{FusterAlexander1971}. While there exist task-optimized RNN models employing attractor dynamics for dynamic coding in memory tasks \citep{Stroud2024}, they lack the resource allocation mechanism central to neural resource theory. Inspired by such considerations, we sought to integrate compressive neural resource allocation with neural attractor dynamics within modern Hopfield networks, creating a novel framework for memory processing that bridges psychological and neuroscientific perspectives.

In this paper, we introduce a \textbf{continuous-time memory} mechanism within modern Hopfield networks, inspired by the continuous attention framework of \citet{martins_CA}. We modify the energy function of \citet{ramsauer2020hopfield}, replacing discrete memory representations with a compressed, continuous alternative. We derive an update rule based on a probability density function (PDF) that links to \cite{martins2022infinite}'s $\infty$-memory transformer. Experiments on synthetic and video datasets show that our approach achieves retrieval performance on par with modern HNs while using a smaller memory, paving the way for more memory-efficient associative models. \footnote{Our code is publicly available at \url{https://github.com/deep-spin/CHM-Net}.}

\section{Hopfield Networks}
\label{sec:background}
Hopfield networks perform associative recall over a set of memory patterns \( \bm{x}_1, \dots, \bm{x}_L \in \mathbb{R}^D \), stored in a memory matrix \( \bm{X} \in \mathbb{R}^{L \times D} \). The network iteratively updates its state \( \bm{q}^{(i)} \) to minimize an energy function, progressively converging toward one of the stored patterns, which serve as stable attractors. Classical Hopfield networks \citep{hopfield1982neural} minimize an energy function given by \( E(\bm{q}) = -\frac{1}{2} \bm{q}^\top \bm{W} \bm{q} \), where \( \bm{q} \) is the query vector and \( \bm{W} = \bm{X}^\top \bm{X} \) is the weight matrix. The state update follows the rule: \( \bm{q}^{(i+1)} = \text{sign}(\bm{W}\bm{q}^{(i)}) \). However, the network’s storage capacity scales only linearly with the input dimensionality, limiting effectiveness for large-scale memory tasks.

To address these limitations, modern Hopfield networks incorporate alternative energy functions that improve storage capacity beyond the classical limit \citep{krotov2016dense,demircigil2017model}. A notable advancement in this domain was proposed by \cite{ramsauer2020hopfield}, who formulated a new energy function for continuous-valued states $\bm{q} \in \mathbb{R}^D$:
\begin{equation}
\label{eq:energyhopfield}
E(\bm{q}) = - \frac{1}{\beta}\log\sum_{i=1}^L \exp(\beta \bm{x}_i^\top \bm{q}) + \frac{1}{2} \|\bm{q}\|^2 + \text{const}.
\end{equation}
A key insight from this formulation is the relationship between modern Hopfield networks and the attention mechanism used in transformers \citep{vaswani2017attention}. Specifically, optimizing (\ref{eq:energyhopfield}) via the concave-convex procedure \citep[CCCP]{yuille2003concave} leads to the update rule:
\begin{equation}
\label{eq:update_rule}
\bm{q}^{(i+1)} = \bm{X}^\top \text{softmax}(\beta \bm{X} \bm{q}^{(i)}),
\end{equation}
By setting the scaling parameter to $\beta = 
\frac{1}{\sqrt{D}}$, this update rule recovers the single-head attention mechanisms in transformers, where identity matrices are used as projection layers. In the next section, we show that modern Hopfield networks can be combined with continuous attention to create more memory-efficient and computationally scalable architectures.

\section{Continuous Attention}
\label{sec:ContinuousAttention}

Traditional attention mechanisms operate on discrete representations, such as words in text or pixels in images. However, many data modalities, such as speech or video, are inherently continuous signals---their discretization with a fixed sampling rate might either be data-inefficient or may fail to exploit their smoothness. Continuous attention \citep{martins_CA, martins2022sparse} addresses this by defining attention over a continuous-time signal \( \bar{\bm{x}}(t) \), treating the input as a representation function evolving smoothly over time. Instead of the probability mass functions having discrete attention weights, this approach leverages a probability density function \( p(t) \), making it well-suited for long or unbounded temporal sequences like time series and audio. The attended output, or context $\bm{c}$, is computed as:
\begin{equation}
   \bm{c} = \mathbb{E}_{p}[\bm{v}(t)] = \int p(t) \bm{v}(t) dt,
\end{equation}
where \( \bm{v}(t) \) denotes the continuous value function (a linear projection of $\bar{\bm{x}}(t)$). For modeling \( p(t) \), \citet{martins2022infinite} use a Gaussian \( \mathcal{N}(t; \mu, \sigma^2) \), where \( \mu \) and \( \sigma^2 \) are input-dependent while \citet{santos2025inftyvideotrainingfreeapproachlong} use uniformly spaced rectangular basis function.

\section{Continuous-Time Memory Hopfield Networks}
\label{eq:cenergyhopfield} 

We assume memories form a continuum, and that sequences of observations \( \bm{X} = [\bm{x}_1^\top, \dots, \bm{x}_L^\top] \in \mathbb{R}^{L \times D} \) are in fact samples from a smooth function  \( \bm{x}(t) \) that exists over a continuous domain. To reconstruct this function, we consider linear combinations of basis functions \( \bm{\psi}(t) \in \mathbb{R}^N \) as
\begin{equation}
\label{eq:con_signal}
    \bar{\bm{x}}(t) = \bm{B}^\top \bm{\psi}(t),
\end{equation}
where \( \bm{B} \in \mathbb{R}^{N \times D} \) denotes the learned coefficients. Usually, the number of basis functions is much smaller than the number of input samples, i.e., \( N \ll L \), ensuring a compressed representation of the memory. The coefficient matrix \( \bm{B} \) is derived through multivariate ridge regression \citep{brown1980adaptive}. %
Each observation in the sequence is assigned a corresponding time point \( t_1, t_2, \dots, t_L \) normalized within the interval \([0,1]\), where the sequence respects the ordering \( t_1 \leq t_2 \leq \dots \leq t_L \) with \( t_\ell \in [0, 1] \). The basis function evaluations at these time points define the design matrix $\bm{F} = [\bm{\psi}(t_1), \dots, \bm{\psi}(t_L)] \in \mathbb{R}^{N \times L}$. The coefficients \( \bm{B} \) are computed such that \( \bm{x}(t_\ell) \approx \bm{x}_\ell \) for each index \( \ell \in \{1, \dots, L\} \), with a regularization parameter \( \lambda > 0 \). This leads to the closed-form solution:
\begin{equation}
    \bm{B}^\top = \bm{X}^\top\bm{F}^\top (\bm{F}\bm{F}^\top + \lambda \bm{I})^{-1}.
\end{equation}

We then define the \textbf{continuous Hopfield energy} using the reconstructed signal \( \bar{\bm{x}}(t) \):
\begin{equation}
\label{eq:energy}
E(\bm{q}) = - \frac{1}{\beta}\log \int_{0}^1 \exp(\beta \bar{\bm{x}} (t)^\top \bm{q}) dt + \frac{1}{2} \|\bm{q}\|^2 + \text{const}.
\end{equation}
This energy function encourages state patterns \( \bm{q} \) to remain close to the linear combination of basis functions \( \bm{\psi}(t) \) represented by the coefficient matrix \( \bm{B} \), according to (\ref{eq:con_signal}). Specifically, \( \bm{\psi}(t) \) form a basis of ``discrete memories,'' and \( \bm{B} \) encodes the weights determining each function’s contribution to \( \bm{q} \). The second term is a quadratic regularization that penalizes large deviations, promoting stability.

The next result states the update rule corresponding to the energy (\ref{eq:energy}).
\begin{proposition}
\label{prop:separation}
Minimizing (\ref{eq:energy}) using the CCCP algorithm \citep{yuille2003concave} leads to the Gibbs expectation update, which is given by:
\begin{equation}
\bm{q}^{(i+1)} = \mathbb{E}_{p(t)}[\bar{\bm{x}}(t)] = \bm{B}^\top \int p(t) \bm{\psi}(t) \, dt,
\end{equation}
where \( p(t)\) is the Gibbs density with temperature \( \beta^{-1} \)
\begin{equation}
p(t) = \frac{\exp(\beta s(t))}{\int \exp(\beta s(t')) \, dt'},
\end{equation}
with the continuous query-key similarity \( s(t) = (\bm{q}^{(i)})^\top \bm{\bar{x}}(t) = (\bm{q}^{(i)})^\top\bm{B}^\top \bm{\psi}(t) \). 
\end{proposition}
The proof is provided in Appendix~\ref{sec:proof_prop_separation}. In our experiments, the integrals in Proposition~\ref{prop:separation} are approximated with the trapezoidal rule. 

\section{Experiments}
\begin{figure*}[t]
\label{fig:archi}
    \centering
    \includegraphics[width=0.95\textwidth]{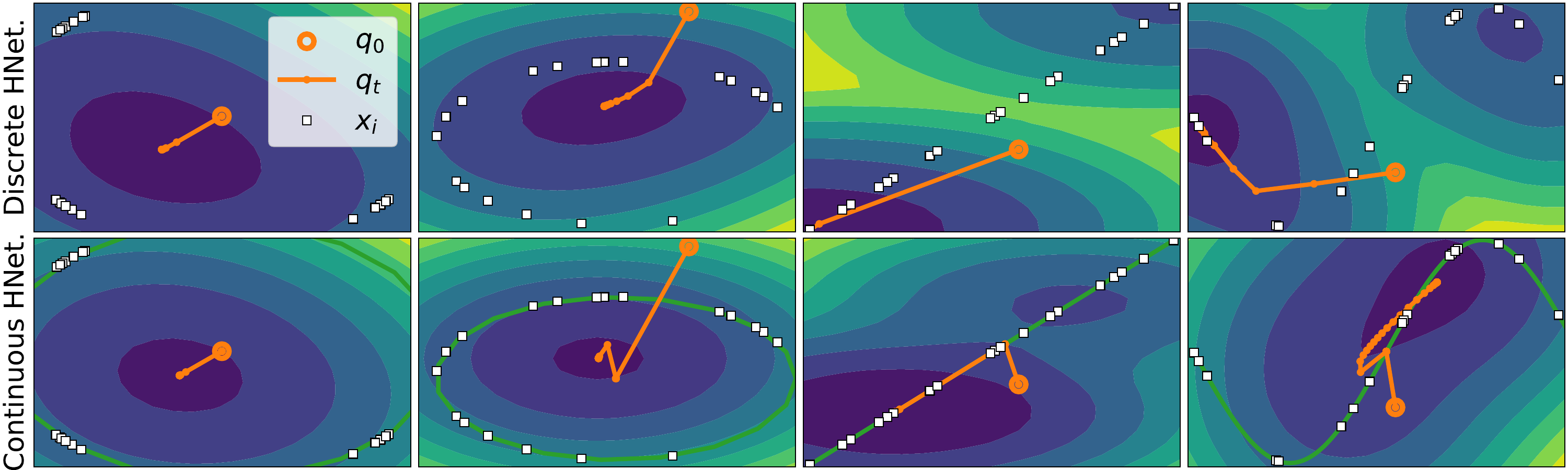}
    \caption{Optimization trajectories and energy contours for Hopfield networks with discrete \textbf{(top)} and continuous memories \textbf{(bottom)}. Green illustrates the continuous function shaped by discrete memory points, while darker shades of blue indicate lower energy regions.}
    \label{fig:contours}
\end{figure*}
\subsection{Hopfield Dynamics and Energy Contours}
In this experiment, we illustrate the optimization trajectories and energy contours for various queries and 20 artificially pattern configurations, sampled from continuous functions, for HNs with both discrete and continuous memories, with \(\beta=1\). For the continuous memory version, we use 10 rectangular basis functions. Figure~\ref{fig:contours} shows comparable retrieval and energy behavior when memory points are sampled from the unit circumference (first and second columns). The final converged point does not correspond to a stored memory due to the dense nature of softmax and Gibbs PDF and the small value of $\beta$. This observation does not hold when the memory function is a line or sinusoid, as the third and fourth columns indicate that for continuous memories, the HN converges to a pattern closer to the initial query than its discrete memory counterpart. The clusters of energy minima, shown in darker blue, are also centered around a group with more points, indicating a more favorable energy landscape for HNs with continuous-time memories. Even when operating with reduced-dimensionality memory representations, continuous memories offer a stable and efficient convergence behavior, making them appealing to replace discrete memories in modern HNs. 

\subsection{Reconstruction of Video Frames}
\begin{figure*}[t]
\label{fig:archi}
    \centering
    \includegraphics[width=0.83\textwidth]{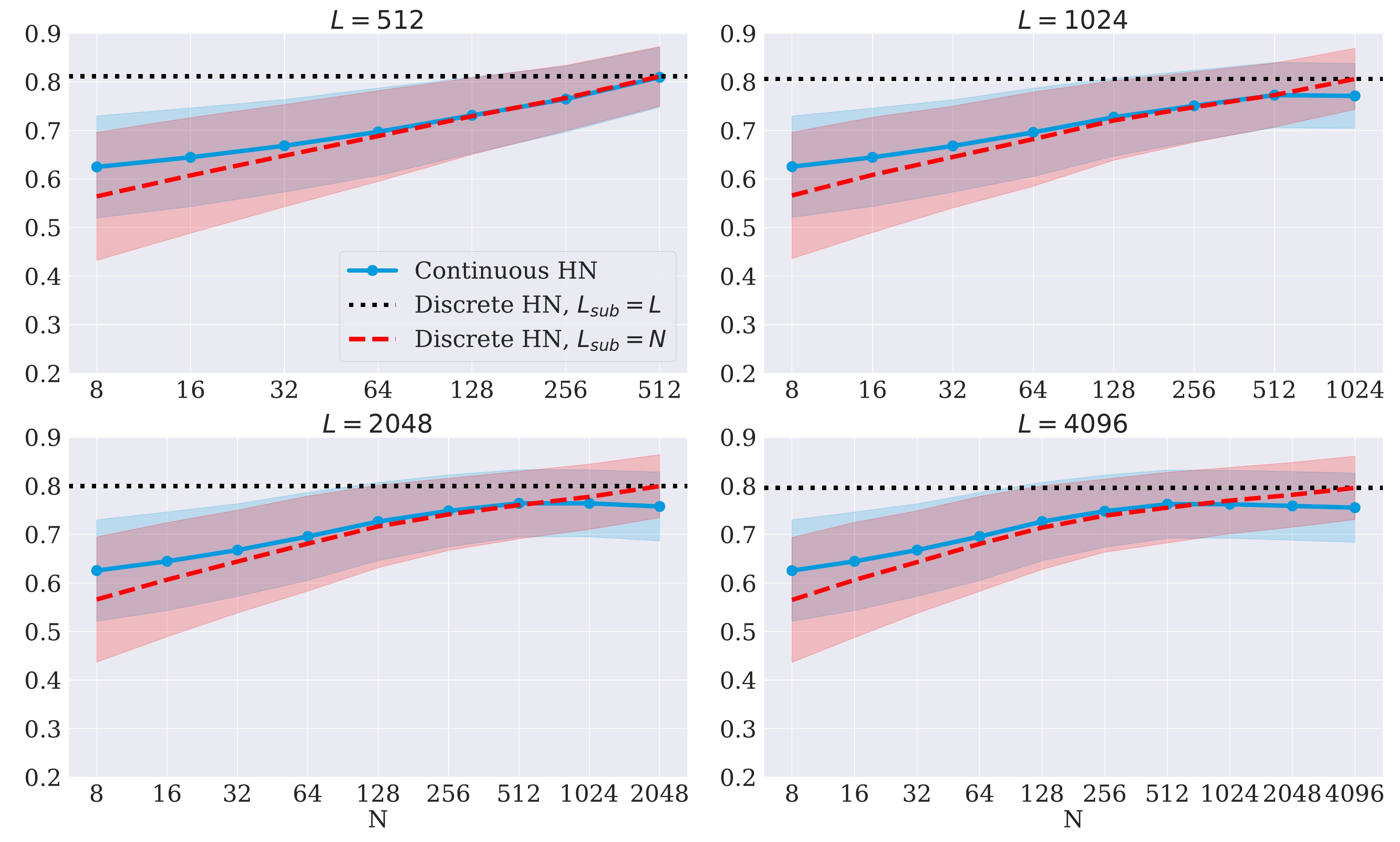}
    \caption{Video retrieval performance across different numbers of basis functions. Plotted are the cosine similarity means and standard deviations across videos.}
    \label{fig:eff1}
\end{figure*}
We now assess the performance of our models in video retrieval using the MovieChat-1K test set \citep{song2023moviechat}, a benchmark of 100 long videos, each averaging 8 minutes. Our focus is on evaluating how HNs with continuous memories perform in retrieving frames from memory sequences of varying lengths, compared to traditional HNs with discrete memories. To do this, we subsample $L$ frames from each video at a resolution of $224 \times 224$ pixels and normalize the pixel values to a range of $[-1, 1]$. Each video is treated as a unique memory. Subsequently, we query both variants of HNs using these memories, with the lower half of each frame masked to $0$.

Figure~\ref{fig:eff1} presents the mean and standard deviation of the cosine similarities between the memories and the retrieved patterns for the HN with continuous memories, across varying values of rectangular basis functions. For the HN with discrete memories, we subsample different numbers of frames, $L_{\text{sub}}$, from the total $L$, ensuring a fair comparison between both models. Our results indicate that for smaller memories, such as $L=512$, the continuous HN consistently outperforms the discrete HN, achieving comparable performance when $N=L$. As $N$ increases, and for larger memory sizes, the continuous HN maintains this trend, although we observe a degradation in performance as $N$ approaches $L$. We hypothesize that this degradation is due to the discrete representation of the queries (\textit{i.e.}, pixel-level representations), which favors the discrete HN, as both the query and memory are discrete. In the following section, we demonstrate that when the reconstruction is performed over continuous domains, such as embeddings, our model shows improvements.   

\subsection{Reconstruction of Video Embeddings}
\begin{figure*}[t]
\label{fig:archi}
    \centering
    \includegraphics[width=0.83\textwidth]{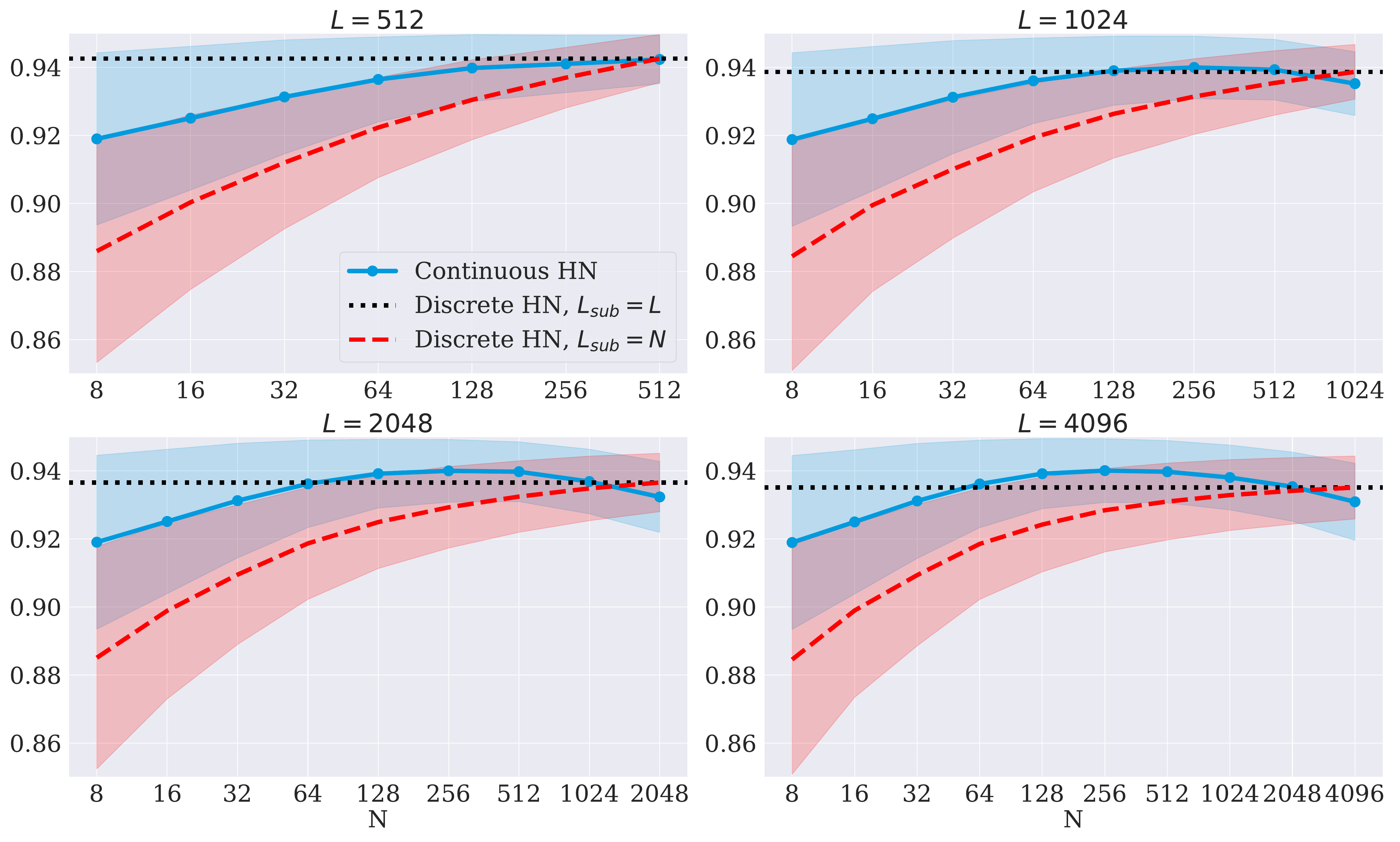}
    \caption{Video embedding retrieval performance across different numbers of basis functions. Plotted are the cosine similarity means and standard deviations across videos.}
    \label{fig:eff2}
\end{figure*}
Next, we present the performance of our method in domains with inherently continuous representations, such as video embeddings. We apply the same pre-processing mentioned earlier; however, we now pass the frames to a visual encoder, specifically EVA-CLIP's ViT-G/14 \citep{fang2022eva}, followed by a Q-former \citep{qformer} that extracts 32 tokens per frame while considering the spatial structure in the frames. To ensure smoothness and continuity, we perform average pooling over the 32 tokens, obtaining $L$ representations. We then add Gaussian noise with $\sigma=5$ to the memories and use them as queries. Figure~\ref{fig:eff2} illustrates the results of embedding reconstruction, emphasizing the consistently superior performance of our HN for larger memories, even surpassing the discrete HN for the full memory $L_{\text{sub}}=L$ but using $N \ll L$, except for $L=512$. This superiority tends to increase as the number of memories \( L \) grows, suggesting that our approach could be a promising direction for improving efficiency in modern HNs. We present ablation studies in Appendix~\ref{app:ablation}.   

\section{Conclusions and Discussion}
We introduced an alternative formulation of Hopfield networks that employs a continuous-time memory, offering a more efficient representation without sacrificing retrieval performance. This approach extends the connection between Hopfield networks and transformer-based architectures such as the $\infty$-former \citep{martins2022infinite} by replacing discrete memories with continuous representations.
Our new energy function introduces an update rule grounded in the continuous attention framework, where the query is iteratively updated according to a Gibbs probability density function over the continuous reconstructed signal.
Memory recall experiments on synthetic and video datasets indicate that our continuous memory formulation achieves retrieval performance on par with discrete Hopfield networks, showcasing its potential for scalable memory-augmented models.

Despite the promising results when $N \ll L$, we observed performance degradation when the number of basis functions approaches the length of the discrete memory. We hypothesize that this stems from the rigid allocation of uniformly spaced rectangular functions, which may not optimally capture the memory’s underlying structure. Future research should explore adaptive mechanisms, such as neural approaches for dynamically learning the widths and centers of basis functions, enabling the model to focus on important regions of the discrete signal. Additionally, replacing the multivariate ridge regression step with a neural network could improve expressiveness and adaptivity in memory modeling. The impact of continuous memories on storage capacity also requires further investigation. Future work will assess the applicability of our approach in real-world problems by exploring the integration of learnable Hopfield layers with continuous memories in practical applications.

\subsubsection*{Acknowledgments}
We thank Sweta Agrawal, Patrick Fernandes, and the SARDINE lab team for helpful discussions. This work was supported by EU's Horizon Europe Research and Innovation Actions (UTTER, contract 101070631), by the project DECOLLAGE (ERC-2022-CoG 101088763), by the Portuguese Recovery and Resilience Plan through project C645008882-00000055 (Center for Responsible AI), and by FCT/MECI through national funds and when applicable co-funded EU funds under UID/50008: Instituto de Telecomunicações.

\bibliography{nfam2025_workshop}
\bibliographystyle{nfam2025_workshop}

\appendix

\section{Proof of Proposition~\ref{prop:separation}}
\label{sec:proof_prop_separation}
The CCCP algorithm works as follows: at the $t$\textsuperscript{th} iteration, it linearizes the concave function $E_{\mathrm{concave}}$ by using a first-order Taylor approximation around $\bm{q}^{(t)}$,
\begin{equation}
    {E}_{\mathrm{concave}}(\bm{q}) \approx \tilde{E}_{\mathrm{concave}}(\bm{q}) := {E}_{\mathrm{concave}}(\bm{q}^{(t)}) + \left(\frac{\partial E_{\mathrm{concave}}(\bm{q}^{(t)})}{\partial \bm{q}}\right)^\top (\bm{q} - \bm{q}^{(t)}).
\end{equation}
Then, it computes a new iterate by solving the convex optimization problem
\begin{equation}
    \bm{q}^{(i+1)} := \arg\min_{\bm{q}} E_{\mathrm{convex}}(\bm{q}) + \tilde{E}_{\mathrm{concave}}(\bm{q}),
\end{equation}
which leads to the equation
\begin{equation}
    \nabla E_\mathrm{convex}(\bm{q}^{(i+1)}) = -\nabla E_\mathrm{concave}(\bm{q}^{(i)}).
\end{equation}
Using the fact that $E_{\mathrm{concave}}(\bm{q})$ and $E_{\mathrm{convex}}(\bm{q})$ are defined as
\begin{align}
    E_{\mathrm{concave}}(\bm{q}) &= - \frac{1}{\beta}\log \int_0^1 \exp(\beta \bar{\bm{x}}(t)^\top \bm{q}) dt, \\
    E_{\mathrm{convex}}(\bm{q}) &= \frac{1}{2} \bm{q}^\top \bm{q},
\end{align}
we compute their gradients:
\begin{align}\label{eq:energy_gradients_new}
    \nabla E_{\mathrm{concave}}(\bm{q}) &= -\frac{1}{\beta} \nabla  \log \int_0^1 \exp(\beta \bar{\bm{x}}(t)^\top \bm{q})\\ 
    \nabla E_{\mathrm{convex}}(\bm{q}) &= \bm{q}.
\end{align}

Using the chain rule for the $E_{\text{concave}}$, we get:
\[
\nabla E_{\text{concave}}(\bm{q}) = - \frac{1}{\beta \int_0^1 \exp(\beta \bar{\bm{x}}(t)^\top \bm{q}) \, dt} \nabla \left( \int_0^1 \exp(\beta \bar{\bm{x}}(t)^\top \bm{q}) \, dt \right).
\]
Next, we compute the gradient of the integral. Since the integral is with respect to \( t \), we differentiate under the integral sign:
\[
\nabla \left( \int_0^1 \exp(\beta \bar{\bm{x}}(t)^\top \bm{q}) \, dt \right) = \int_0^1 \beta \bar{\bm{x}}(t) \exp(\beta \bar{\bm{x}}(t)^\top \bm{q}) \, dt.
\]
Therefore, the gradient of \( E_{\text{concave}}(\bm{q}) \) becomes:
\[
\nabla E_{\text{concave}}(\bm{q}) =  -\frac{\int_0^1 \bar{\bm{x}}(t) \exp(\beta \bar{\bm{x}}(t)^\top \bm{q}) \, dt}{\int_0^1 \exp(\beta \bar{\bm{x}}(t)^\top  \bm{q}) \, dt} = -\int_0^1 p(t) \bar{\bm{x}} (t) dt,
\]
where $p(t)$ is the Gibbs probability density function. Now, recognizing that the above expression represents the expectation with respect to the distribution \( p(t) \), we write:
\[
\nabla E_{\text{concave}}(\bm{q}) = -\mathbb{E}_{p(t)}[\beta \bar{\bm{x}}(t)].
\]
Thus, the update equation is given by
\begin{equation}\label{eq:updates_hfy_new}
    \bm{q}^{(i+1)} = \mathbb{E}_{p(t)} [\bar{\bm{x}}(t)],
\end{equation}
which corresponds to the Gibbs expectation update.

\section{Ablation Studies}
\begin{figure*}[t]
\label{fig:archi}
    \centering
    \includegraphics[width=1\textwidth]{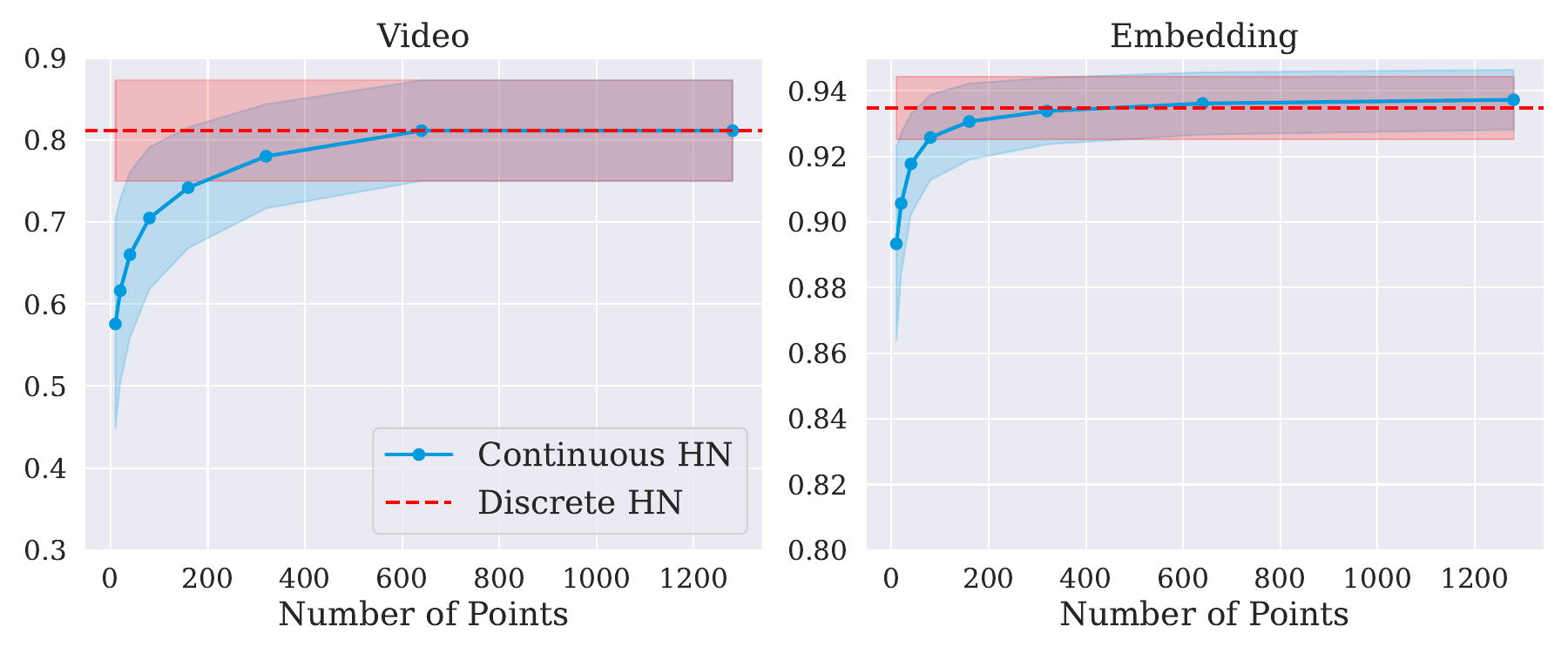}
    \caption{Performance on video and embedding data across different numbers of sampling points used to approximate the integrals of our framework.}
    \label{fig:points}
\end{figure*}

We present in Figure~\ref{fig:points} the average cosine similarities as a function of the number of points used to approximate the integrals from Proposition~\ref{prop:separation}. For the video dataset with 50\% masked frames, we use $L=512$ and $N=512$. For embedding reconstruction, we set $L=2048$ and $N=1024$ with $\sigma = 5$. The results demonstrate that 500 sampling points are sufficient for the approximation, where the Hopfield network with continuous memories performs comparably to the modern Hopfield network.
\label{app:ablation}

\end{document}